\documentclass[10pt]{article}

\usepackage[utf8]{inputenc}
\usepackage[T1]{fontenc}
\usepackage[a4paper,margin=1.25in]{geometry}
\usepackage{setspace}
\usepackage{mathpazo}
\usepackage{amsmath,amssymb,amsfonts,amsthm,mathtools}
\usepackage{graphicx}
\usepackage{booktabs}
\usepackage{array,tabularx}
\usepackage{enumitem}
\usepackage{xcolor}
\usepackage[colorlinks=true,linkcolor=blue,citecolor=blue,urlcolor=blue]{hyperref}
\usepackage[pageref]{backref}

\numberwithin{equation}{section}
\setlist[itemize]{nosep,leftmargin=*}
\setlist[enumerate]{nosep,leftmargin=*}
\newcolumntype{Y}{>{\raggedright\arraybackslash}X}

\theoremstyle{definition}

\theoremstyle{remark}

\newcommand{\E}{\mathbb{E}}
\newcommand{\N}{\mathbb{N}}
\newcommand{\R}{\mathbb{R}}
\newcommand{\Z}{\mathbb{Z}}
\newcommand{\Wcal}{\mathcal{W}}

\newcommand{\Vcal}{\mathcal{V}}

\renewcommand*{\backrefalt}[4]{%
\ifcase #1 %
(Not cited)%
\or
(Cited on p.~#2)%
\else
(Cited on pp.~#2)%
\fi
}

\title{AI Tokenomics: The Economics of Tokens, Computation, and Pricing in Foundation Models}
\author{%
Quanyan Zhu\thanks{Department of Electrical and Computer Engineering, New York University Tandon School of Engineering, 370 Jay Street, Room 1004, Brooklyn, NY 11201, USA. Email: \href{mailto:qz494@nyu.edu}{qz494@nyu.edu}; phone: +1 646-997-3371.}
}
\date{}

\begin{document}
\maketitle

\begin{abstract}
Tokens have become the practical accounting unit for modern foundation
model services, linking information processing, computation, memory use,
energy expenditure, pricing, and economic value. This paper develops a
framework for AI tokenomics: the study of how tokens are generated,
consumed, priced, allocated, and optimized across AI systems. We connect
token-level technical costs to workflow-level production functions,
enterprise resource allocation, measurement and instrumentation methods,
and emerging market-design questions. The framework shows that token
expenditure and economic value are distinct: value depends on marginal
productivity, workflow position, hidden reasoning activity, risk, and
downstream propagation effects. The paper concludes by identifying open
research directions in hidden-token measurement, empirical calibration,
token productivity, dynamic allocation, and token-based markets.
\end{abstract}


\section{Introduction}

The rapid commercialization of foundation models has created a new economic
unit for artificial intelligence: the token. In its original technical
sense, a token is a discrete representation used to process language and
other data in transformer architectures
\cite{vaswani2017attention,radford2019language,brown2020language}. In its
emerging economic sense, however, the token has become the unit through
which foundation-model services meter, price, allocate, and govern
computational intelligence. Every prompt, retrieval operation, reasoning
step, tool invocation, memory access, and generated response is ultimately
converted into token consumption. Token usage therefore links information
processing to computation, latency, memory, energy expenditure, and
monetary cost. This dual role is unusual: the same object that determines
how a model parses and generates information also determines how users are
charged and how organizations reason about capacity, utilization, and
return on investment.

This shift matters now because AI is moving from experimental deployment to
large-scale enterprise infrastructure. Recent industry analyses argue that
AI is becoming one of the fastest-growing categories of enterprise
technology expenditure and that organizations must manage this expenditure
at the level of token consumption \cite{deloitte2026tokenomics}. Unlike
traditional software systems whose costs are often governed by licenses,
seats, or virtual machines, foundation-model deployments exhibit volatile
and nonlinear usage patterns. Token demand depends on prompt complexity,
context size, reasoning depth, retrieval design, tool use, workflow
architecture, and autonomous agent interactions. Small changes in a prompt,
retrieval policy, or agentic workflow can therefore create large changes in
cost when repeated across many users, tasks, or automated agents. As a
result, conventional total-cost-of-ownership and cloud-accounting
frameworks are increasingly insufficient for managing AI systems at scale.
An enterprise no longer needs only to ask how many servers, licenses, or
cloud instances it has purchased. It must also ask how many tokens different
business processes consume, which workflows produce value per token, where
hidden reasoning or retrieval costs arise, and when a deployment should
shift across API, SaaS, or self-hosted infrastructure.

Token-based pricing reinforces this transformation. Historically,
computational resources such as processor cycles, memory, storage, and
network bandwidth were measured and managed through distinct accounting
mechanisms. Contemporary foundation-model services instead expose a unified
economic interface in which heterogeneous computational resources are
abstracted into token usage. Whether an organization relies on proprietary
APIs, self-hosted models, retrieval-augmented generation (RAG), or
autonomous AI agents, resource utilization is increasingly expressed in
tokens. The widespread adoption of token-based pricing by OpenAI,
Anthropic, Google, xAI, and other providers makes tokens a common
accounting interface for computational intelligence
\cite{openai_pricing_2026,anthropic_pricing_2026,
google_gemini_pricing_2026,xai_pricing_2026}.

\subsection{From Token Accounting to AI Tokenomics}

Current industry practice often treats this interface primarily as a
problem of accounting, cost management, infrastructure optimization, and
Financial Operations (FinOps) for AI. FinOps is a management practice
concerned with making cloud and AI spending visible, accountable, and
controllable. In an AI setting, it helps organizations track token
expenditures, forecast budgets, assign costs to teams or applications, and
reduce waste. These
activities are necessary for operational control, but they do not exhaust
the economic questions created by tokenized AI systems. Once tokens become
the unit through which AI activity is measured, organizations must also
determine how tokens should be valued, allocated, priced, forecast,
governed, and optimized across complex AI ecosystems. This broader set of
questions motivates the field that we refer to as \emph{AI Tokenomics}. AI
tokenomics studies the generation, consumption, pricing, valuation,
allocation, optimization, and governance of tokens within AI ecosystems. Its
central premise is that tokens should not be viewed merely as artifacts of
tokenization schemes such as Byte Pair Encoding (BPE) or SentencePiece
\cite{sennrich2016bpe,kudo2018sentencepiece}. Rather, they are economic
resources that mediate the relationship between computational effort,
workflow performance, organizational value, and market structure.
In this sense, AI tokenomics begins where token accounting and FinOps end.
Accounting records token use after it occurs, and FinOps helps manage the
resulting expenditure. Tokenomics asks a broader design question: how should
token use be predicted, priced, constrained, and allocated before and during
operation so that tokens generate value rather than merely cost? This
requires studying the marginal productivity of tokens, tradeoffs among
models and workflows, demand created by users and agents, risk propagation
through workflow dependencies, and allocation rules for scarce computational
capacity.

From this perspective, the increasing adoption of foundation models is
transforming AI from a purely computational technology into an economic
system whose fundamental unit of activity is the token. Tokens determine
how computational resources are consumed, how providers generate revenue,
how organizations incur costs, how workflows allocate scarce resources, and
how value is created and distributed across AI ecosystems. They play a role
analogous to energy in industrial systems or bandwidth in communication
networks: they are a resource through which productive capacity is
generated, delivered, and consumed. Understanding the economics of this
resource is therefore essential for the efficient design, operation, and
governance of large-scale AI systems.

\subsection{Research Scope}

Several structural features distinguish token economics from standard
allocation problems. First, token consumption exhibits strong
nonlinearities. Empirical scaling-law studies show that model performance
depends systematically on model size, data volume, and computational
expenditure \cite{kaplan2020scaling,hoffmann2022training}. Additional
context, retrieval, reasoning, or tool usage may improve performance, but
the associated gains often display diminishing returns. Second, token
demand is endogenous and adaptive: it depends not only on task
characteristics but also on prompt design, workflow structure, model
behavior, and autonomous decision making. Third, token consumption may
contain hidden components associated with internal reasoning,
chain-of-thought computation, memory operations, and planning processes that
are not directly observable to users. Fourth, token prices differ across
providers, modalities, context windows, caching regimes, and deployment
models, creating heterogeneous economic landscapes for similar tasks.

The complexity of token economics increases further in enterprise settings,
where modern AI systems rarely operate as isolated model calls. They are
increasingly assembled as interconnected workflows involving retrieval
systems, reasoning modules, planning engines, monitoring components,
tool-using agents, and autonomous decision-support systems
\cite{yao2023react,schick2023toolformer}. Token allocations made at one
stage of a workflow can influence the quality, reliability, risk, and value
generated by downstream stages. Consequently, token allocation becomes a
networked optimization problem in which local decisions have system-wide
effects. At the same time, token markets raise questions about pricing,
incentives, access, and governance as AI services become more deeply
embedded in organizational and economic life.
The relevant unit of analysis therefore changes with scale. At the
task level, tokenomics concerns how difficulty, context, uncertainty, and
output requirements generate token demand. At the workflow level, it
concerns how tokens should be allocated across interdependent stages whose
qualities complement or substitute for one another. At the enterprise level,
it concerns how token budgets should be governed under uncertainty, risk,
and organizational objectives. At the market level, it concerns how token
prices, contracts, and allocation mechanisms shape access to computational
intelligence.

These observations motivate several fundamental research questions. At the
technical level, how do tokens map to computation, memory, latency, and
energy consumption? At the task level, how do difficulty, context size, and
uncertainty shape token demand? At the enterprise level, how should
organizations allocate finite token budgets across competing workflows, and
how should token productivity and economic value be measured? At the market
level, what determines token prices across providers and deployment
environments, and what forms of pricing mechanisms, allocation rules, and
governance structures may emerge as AI systems become increasingly
token-centric?

This paper develops a framework for addressing these questions by integrating
technical foundations, economic analysis, enterprise resource allocation,
measurement methodologies, and market design into a unified theory of AI
tokenomics. The analysis proceeds from token-level technical costs to
task-dependent token demand, provider pricing, workflow-level allocation,
instrumentation, and market design. Throughout, tokens are treated not
merely as accounting units but as allocable economic resources whose
efficient use is fundamental to AI systems.

\subsection{Contributions}

Existing work provides important but fragmented foundations for this
problem. Tokenization research explains how raw inputs are converted into
discrete model inputs \cite{sennrich2016bpe,kudo2018sentencepiece}; scaling
laws relate model performance to data, parameters, and compute
\cite{kaplan2020scaling,hoffmann2022training}; provider documentation and
pricing pages describe how tokens are billed
\cite{openai_pricing_2026,anthropic_pricing_2026,
google_gemini_pricing_2026,xai_pricing_2026}; and
FinOps practice focuses on tracking and controlling AI expenditure. These
perspectives do not yet provide a unified theory of tokens as economic
resources. In particular, existing treatments often view tokens either as
linguistic units inside a model or as billing units on an invoice, but not
as scarce computational resources whose allocation determines workflow
quality, enterprise value, risk exposure, and market structure.

This paper addresses that gap in six ways. First, it formalizes AI
tokenomics as the study of how tokens are generated, consumed, priced,
valued, allocated, optimized, and governed within AI ecosystems. This
definition moves beyond token accounting by treating tokens as resources
that can be assigned, substituted, conserved, and priced. Second, the paper
develops a technical foundation that links token categories to computation,
memory utilization, hidden reasoning, energy consumption, and inference
cost. Third, it introduces task-level token-demand and token-production
models in which difficulty, context size, and uncertainty determine token
requirements, thereby creating a basis for empirical calibration across
heterogeneous workloads. Fourth, it analyzes contemporary provider pricing
structures, instrumentation methods, and measurement challenges, including
input-output price asymmetries, caching, subscriptions, enterprise
agreements, hidden reasoning tokens, and energy profiling. Fifth, it
develops a workflow-level allocation model in which interdependent AI
workflows form a network, workflow quality depends on token allocations,
and optimal token use is characterized through marginal productivity,
shadow prices, downstream propagation, and risk-aware allocation. Sixth, it
extends the analysis to market and mechanism design by studying dynamic
pricing, tokenized compute assets, contract design, and multi-agent token
economies. Numerical case studies then illustrate how token expenditures,
workflow quality, and economic value interact in realistic AI deployments.

\subsection{Organization of the Paper}

Section~\ref{sec:technical} develops the technical anatomy of tokens and
models task-level token consumption. Section~\ref{sec:pricing} examines
token pricing, provider cost structures, instrumentation, and computational
profiling. Section~\ref{sec:markets} studies market-design questions for
tokenized compute, multi-agent token economies, and token-allocation
mechanisms. Section~\ref{sec:workflow} then develops one concrete mechanism
from that taxonomy: an enterprise-internal workflow allocation model with
networked resource allocation and risk-aware optimization.
Section~\ref{sec:case-studies} presents case
studies illustrating how token expenditure, workflow quality, and economic
value interact in practical settings. Section~\ref{sec:conclusion}
concludes with open research directions.

\section{Technical Anatomy of Tokens and Task-Dependent Token Consumption}
\label{sec:technical}

This section provides the technical foundation for the tokenomic framework
developed in the rest of the paper. It first defines tokens as discrete
computational objects produced by a tokenizer, then distinguishes the major
categories of token usage that appear in deployed AI systems. The section
then connects token counts to computation, energy, pricing, and task-level
demand, thereby establishing the notation and modeling assumptions used in
the subsequent economic and workflow analyses.

\subsection{Tokens as Computational and Economic Units}

Large Language Models (LLMs) operate on discrete units of information known as
\emph{tokens}. A token constitutes the fundamental unit through which textual,
audio, image, or other forms of data are represented and processed by modern
foundation models. Beyond their linguistic role, tokens have emerged as the
primary unit of computation, accounting, and pricing in contemporary AI
services. Consequently, understanding the anatomy of tokens is essential for
both the engineering analysis of large-scale AI systems and the development of
a rigorous theory of AI tokenomics.

Throughout the paper, $\R_+=[0,\infty)$, $\R_{++}=(0,\infty)$,
$\Z_{\ge0}=\{0,1,2,\ldots\}$, and $\N=\{1,2,\ldots\}$. Formally, let
$\mathcal X$ denote the raw input space and let $\Vcal$ denote
the token vocabulary. For text models, $\mathcal X$ may be taken as the set of
finite strings over an alphabet; for multimodal models, it may be enlarged to
finite records containing text, image, audio, or tool-state fields. Let
$\Vcal^{*}=\bigcup_{n\ge0}\Vcal^n$ denote the set of finite token sequences. A
tokenizer is a mapping $\tau:\mathcal X\rightarrow\Vcal^{*}$. For
$x\in\mathcal X$, write $\tau(x)=(t_1,\ldots,t_n)$ with
$t_i\in\Vcal$ and $n=|\tau(x)|\in\Z_{\ge0}$.
The precise decomposition depends on the tokenization scheme employed, such as
Byte Pair Encoding (BPE), WordPiece, or SentencePiece
\cite{sennrich2016bpe,kudo2018sentencepiece}. Modern foundation models rely
on such tokenization procedures to transform raw inputs into discrete
representations suitable for transformer-based processing
\cite{brown2020language}.

In English text, one token typically corresponds to approximately three to
four characters, or roughly three-quarters of a word, although substantial
variation exists across languages, writing systems, and application domains.
From the perspective of inference, tokens represent the atomic objects upon
which transformer architectures operate. Every attention computation, memory
access, and prediction step is ultimately conditioned on token
representations. Consequently, the total computational burden of an AI service
is closely tied to the number of tokens processed during execution.

Beyond their computational role, tokens have acquired economic significance.
Commercial AI platforms meter usage, allocate resources, and determine pricing
primarily through token counts. As a result, tokens simultaneously function as
units of information representation, computation, and economic exchange,
forming the fundamental building blocks of modern AI tokenomics.

\subsection{Categories of Tokens}

Although token accounting is often reported as a single aggregate quantity,
multiple conceptually distinct categories of tokens exist within modern AI
systems. Unless otherwise stated, realized token counts belong to
$\Z_{\ge0}$; in optimization problems below, these counts are relaxed to
$\R_+$ to permit continuous marginal analysis. Input tokens, denoted by
$T_I\in\Z_{\ge0}$, correspond to user prompts, system instructions, and
task-specific information supplied before inference begins.
Context tokens, denoted by $T_C\in\Z_{\ge0}$, represent conversational history or
persistent memory retained within the model's context window. Retrieval-based
systems introduce an additional category of tokens, denoted by
$T_R\in\Z_{\ge0}$,
corresponding to external documents incorporated through
Retrieval-Augmented Generation (RAG) \cite{lewis2020rag}.

The model subsequently generates output tokens, denoted by
$T_O\in\Z_{\ge0}$, which
constitute the visible response returned to the user. More subtly, advanced
reasoning models frequently generate intermediate computational traces
associated with chain-of-thought reasoning and latent deliberation processes
\cite{wei2022chain,wang2023selfconsistency}. These hidden reasoning tokens,
denoted by $T_H\in\Z_{\ge0}$, are sometimes referred to as \emph{thinking tokens} or
\emph{internal chain-of-thought tokens}. Although typically inaccessible to
users, they may contribute substantially to computational cost, memory
consumption, and energy expenditure.

The total token footprint associated with a task may therefore be expressed as

\begin{equation}
T_{\mathrm{tot}}
=
T_I + T_C + T_R + T_O + T_H.
\label{eq:token-footprint}
\end{equation}

\noindent This decomposition highlights an important distinction between observable and
unobservable token expenditures. Whereas users generally observe only input
and output tokens, infrastructure providers must provision resources for the
entire quantity $T_{\mathrm{tot}}$. Consequently, token accounting should be
viewed as a system-level concept rather than merely a measure of visible user
interaction.

\subsection{Tokens, Computation, and Resource Consumption}

The economic significance of tokens derives from their close relationship with
computational complexity. Consider a transformer architecture containing $L$
layers and model width $m$, where $L,m\in\N$ \cite{vaswani2017attention}. Each token
passing through the network undergoes repeated applications of self-attention
and feedforward transformations. Ignoring constant factors, the computational burden of processing a sequence
of length $T$ scales approximately as $O(Lm^{2}T)$. A more refined analysis
reveals that self-attention introduces an additional dependence on sequence
length. Since each token attends to all preceding
tokens, attention operations scale approximately as $O(T^{2})$
\cite{vaswani2017attention}. Consequently, extending context windows from
$10^{4}$ to $10^{5}$ tokens can increase computational requirements by orders
of magnitude, motivating the development of sparse-attention architectures,
FlashAttention, and memory-efficient inference mechanisms
\cite{dao2022flashattention}.

The relationship between token count and computational effort may be
summarized by

\begin{equation}
F(T)\approx \kappa L m^{2} T,
\label{eq:flops-token}
\end{equation}

\noindent where $F:\R_+\rightarrow\R_+$ denotes the total number of floating-point
operations (FLOPs) required for inference and $\kappa>0$ is an
architecture-dependent constant. Although simplified, this expression emphasizes that every additional token
carries a measurable computational cost.
This observation underlies contemporary scaling-law analyses, which reveal
systematic relationships among model size, data volume, computational
expenditure, and downstream performance
\cite{kaplan2020scaling,hoffmann2022training}. From this perspective, tokens
represent a practical proxy for computational effort and therefore provide a
natural unit for economic accounting.

Tokens also determine memory utilization. During autoregressive inference,
transformers maintain a key-value (KV) cache storing intermediate
representations associated with previously processed tokens. If the model
contains $L$ layers and model width $m$, the memory function
$M:\R_+\rightarrow\R_+$ satisfies $M(T)\propto LmT$ in the simplified
linear approximation. This linear
dependence on context length explains why long-context models
require substantial GPU memory resources and why context management has become
a critical design challenge in large-scale AI systems.

Because each token triggers computation and memory accesses, token processing
is directly associated with both energy consumption and latency. Let
$\bar e\in\R_+$ denote the average energy expenditure per token and let
$\bar\ell\in\R_+$ denote the average latency per token. The model-based
energy-token mapping for a workload containing $T$ tokens may then be
approximated by $\mathcal E_{\mathrm{tok}}(T)=\bar e T$, while total latency
is approximately $\ell_{\mathrm{tot}}=\bar\ell T$. These relationships motivate metrics such as
\emph{Joules per token} and
\emph{milliseconds per token}, which provide hardware-independent measures of
inference efficiency.

\subsection{Tokens as Economic Objects}

The emergence of commercial AI services has transformed tokens from purely
technical entities into economic objects. Most providers charge separately for
input and output tokens. If $p_I,p_O\in\R_+$ denote the respective prices per
token, the cost function $C:\R_+^4\rightarrow\R_+$ for a single interaction
may be represented as
$C=p_I(T_I+T_C+T_R)+p_OT_O$. This pricing model reflects the fact that
generating output tokens generally requires more computation than processing
input tokens. More fundamentally, however, token-based pricing reveals a
conceptual shift in the role of
tokens. Tokens are no longer merely units of information representation; they
function simultaneously as units of computation, resource allocation, and
economic exchange.
In this sense, tokens play a role analogous to kilowatt-hours in electricity
markets or packet counts in communication networks. They provide a common
currency through which computational effort, infrastructure utilization, and
economic value may be quantified. This dual technical-economic character forms
the foundation of AI tokenomics.

\subsection{Task-Dependent Token Consumption}

While the preceding discussion explains why tokens possess computational and
economic significance, it does not explain why different AI tasks consume
dramatically different numbers of tokens. Token consumption is not determined
solely by prompt length. Rather, it emerges from the interaction among task
complexity, contextual requirements, and uncertainty in the reasoning
process. Prompt construction, context-window management, and per-call usage
instrumentation therefore enter token demand directly
\cite{openai_promptguide2026,anthropic_context2026,langchain_openai_tokens2026}.

From an economic perspective, token usage may be viewed as a derived demand
for computational reasoning. Just as electricity demand depends on the
activities performed by a factory, token demand depends on the informational
and computational requirements of the underlying task. A simple fact lookup
or classification query may require only a few hundred to a few thousand
tokens, whereas long-document summarization may require $10^4$--$10^5$ or
more tokens. Code generation often has moderate input demand but large
output demand, mathematical proof and planning tasks can consume substantial
hidden reasoning tokens, and agentic systems amplify usage through repeated
model calls, tool invocations, verification loops, and subtask decomposition
\cite{wei2022chain,wang2023selfconsistency,yao2023react,schick2023toolformer,bai2026agenttokenconsumption}.

Let $\mathcal J=\{1,\ldots,J\}$, with $J\in\N$, denote a finite set of task
classes. For each $j\in\mathcal J$, let
$\widehat T_j:\Omega\rightarrow\Z_{\ge0}$ be the realized total token
footprint of an execution of task class $j$, where $\Omega$ is the underlying
probability space capturing model stochasticity, prompt variation, retrieval
variation, and tool-use paths. Define the task descriptor
$\theta_j=(d_j,c_j,u_j)\in\Theta=\mathcal D\times\mathcal C\times\mathcal U$,
where $\mathcal D,\mathcal U\subseteq\R_{++}$ are normalized dimensionless
scales for task difficulty and uncertainty or creativity, respectively, and
$\mathcal C\subseteq\R_{++}$ is a context-size domain measured in tokens or
normalized context units. The expected token-demand mapping
$f:\Theta\rightarrow\R_+$ is defined by

\begin{equation}
T_j=\E[\widehat T_j\mid d_j,c_j,u_j]=f(d_j,c_j,u_j),
\label{eq:token-demand}
\end{equation}

\noindent where $T_j$ is the continuous relaxation of expected token demand for task
class $j$.

The variable $d_j$ measures the computational complexity of the reasoning
process, including the number and difficulty of intermediate steps required
to solve the task. The variable $c_j$ captures the quantity of information
that must be processed, including prompt text, retrieved evidence,
conversation history, and tool outputs. The variable $u_j$ represents
uncertainty, ambiguity, or creativity requirements, and therefore captures
the breadth of the solution space explored during inference. Two tasks with
identical prompt lengths may therefore consume vastly different numbers of
tokens if they differ substantially in difficulty or uncertainty.

A useful parametric representation for positive descriptors is
$T_j=a d_j^{\alpha}c_j^{\beta}u_j^{\gamma}$, where
$a\in\R_{++}$ and $\alpha,\beta,\gamma\in\R_{++}$. The exponents
$\alpha$, $\beta$, and $\gamma$ are elasticities of token consumption with
respect to task difficulty, context size, and uncertainty. Precise
calibration requires empirical logs of token consumption across diverse
tasks. As an illustrative approximation, context size often enters nearly
linearly, so $\beta\approx1$, whereas uncertainty or creativity may have a
stronger and potentially super-linear effect, such as $\gamma>1$.

Importantly, this task-level production function serves as the microeconomic
foundation for the workflow-level allocation models developed later in the
paper. Whereas the present formulation explains how task characteristics
generate token demand, the enterprise tokenomics framework subsequently
examines how finite token budgets should be allocated across interconnected
workflows to maximize organizational value.

\subsection{Empirical Profiles of Token Demand}

The heterogeneity of token consumption becomes evident when examining common
AI workloads. The ranges in Table~\ref{tab:token_profiles} are representative
engineering profiles rather than universal constants; they depend on model,
tokenizer, context-window policy, retrieval depth, prompt design, and
orchestration strategy. They are nevertheless useful for illustrating the
orders-of-magnitude differences across task classes and for motivating
empirical token logging at the application level
\cite{anthropic_context2026,langchain_openai_tokens2026,bai2026agenttokenconsumption}.

\begin{table}[ht]
\centering
\scriptsize
\caption{Representative token-consumption profiles across common AI workloads.}
\label{tab:token_profiles}
\begin{tabularx}{\textwidth}{@{}p{1.9cm}p{2.1cm}p{2.15cm}p{1.65cm}p{2.05cm}Y@{}}
\toprule
\textbf{Task type} &
\textbf{Example} &
\textbf{Input/context} &
\textbf{Output} &
\textbf{Hidden/reasoning} &
\textbf{Drivers} \\
\midrule
Chat Q\&A &
FAQ or classification &
$50$--$1{,}000$ &
$50$--$1{,}000$ &
Few &
Low $d_j$, low $c_j$, low $u_j$ \\
Document summarization &
Multi-page summary &
$10^4$--$10^5+$ &
$100$--$2{,}000$ &
Moderate &
Moderate $d_j$ and high $c_j$ \\
Code generation &
Function implementation &
$20$--$2{,}000$ &
$500$--$10{,}000$ &
Moderate &
Moderate $d_j$ and variable $c_j$ \\
Math proof or analysis &
Theorem or problem solution &
$50$--$2{,}000$ &
$300$--$5{,}000$ &
Very high &
Very high $d_j$ and high $u_j$ \\
Planning or RAG workflow &
Retrieval and planning &
$1{,}000$--$10{,}000$ &
$200$--$1{,}000$ &
High &
High $d_j$, high $c_j$, high $u_j$ \\
Multi-agent task &
Subtask orchestration &
Varies across turns and tools &
Varies &
Compounded across calls &
Recursive decomposition across subtasks \\
\bottomrule
\end{tabularx}
\end{table}

Typical total token scales implied by these profiles range from roughly
$200$--$2{,}000$ tokens for chatbot-sized simple queries, to
$2{,}000$--$12{,}000$ tokens for retrieval-augmented question answering, to
$10{,}000$--$100{,}000+$ tokens for multi-page summarization. Agentic or
multi-step workflows can reach $15{,}000$--$1{,}000{,}000+$ tokens because
each planning, retrieval, tool-use, and verification step may induce another
model call. Recent evidence on agentic coding tasks similarly finds that
agentic executions can be orders of magnitude more token intensive than
ordinary code chat or single-step reasoning \cite{bai2026agenttokenconsumption}.

Several observations emerge from Table~\ref{tab:token_profiles}. First,
contextual information frequently dominates token demand in enterprise
applications. Second, reasoning-intensive tasks often consume substantial
numbers of hidden tokens that may not be directly observable to users.
Third, autonomous and multi-agent systems exhibit token consumption that
grows through recursive decomposition of tasks into subtasks, producing
token expenditures several orders of magnitude larger than those associated
with traditional chatbot interactions.

\section{Token Pricing, Measurement, and Instrumentation}
\label{sec:pricing}

This section translates the technical accounting of tokens into the economic
and operational language used by AI providers and enterprise users. The main
objective is to show how token counts become prices, invoices, budgets, and
measurement signals, while also identifying the parts of inference cost that
remain difficult to observe. The discussion begins with provider pricing and
then turns to subscription models, instrumentation, hidden reasoning, and
energy profiling.

\subsection{Token Pricing as an Accounting Interface}

The commercialization of foundation models has led to the widespread adoption
of token-based pricing as the dominant mechanism for allocating computational
resources. Because tokens serve as a direct proxy for inference workload,
memory utilization, and infrastructure consumption, they provide a natural
basis for metering AI services. Nearly all major providers, including OpenAI,
Anthropic, Google, and xAI, price access to their models according to the
number of tokens processed during inference
\cite{openai_pricing_2026,anthropic_pricing_2026,
google_gemini_pricing_2026,xai_pricing_2026}.

From the provider's perspective, token pricing offers several advantages.
First, it aligns revenue with resource consumption, thereby allowing costs to
scale with demand. Second, token accounting provides a model-agnostic unit of
measurement that remains applicable across diverse architectures and
applications. Third, it enables granular pricing structures that distinguish
between different categories of computational workload, such as prompt
processing, response generation, caching, and long-context inference. Despite
the apparent simplicity of token-based billing, substantial variation
exists across providers. Differences arise in input and output token rates,
treatment of cached tokens, handling of reasoning tokens, and support for
batch-processing discounts. These distinctions reflect both underlying
differences in computational costs and strategic considerations regarding
market positioning and customer acquisition. Accordingly, a rigorous theory of
AI tokenomics must treat pricing and measurement together. Commercial AI
platforms routinely report input and
output token counts, yet many aspects of token utilization remain opaque. In
particular, hidden reasoning processes, internal memory operations, and
hardware-level resource consumption are typically inaccessible to end users.
Consequently, pricing is not merely a billing convention; it is also an
observable interface through which users infer otherwise hidden computational
costs.

\subsection{Provider Pricing Structures and Price Asymmetries}

Table~\ref{tab:foundation_models} gives a representative snapshot of token
pricing and technical characteristics among major foundation-model providers
as of June 2026. The figures should be interpreted as illustrative rather
than permanent rates, since pricing evolves rapidly as model capabilities
improve, context windows expand, and providers introduce new service tiers
\cite{openai_pricing_2026,anthropic_pricing_2026,
google_gemini_pricing_2026,xai_pricing_2026}. Official provider pages remain
the authoritative source for billing, while third-party dashboards provide a
useful cross-provider comparison layer for scenario analysis and models whose
commercial terms are distributed across platform-specific pages
\cite{benchlm_2026,costgoat_2026,deepseek_pricing_2026,qwen_pricing_2026,
glm_pricing_2026,meta_llama_2026,mistral_pricing_2026}.

\begin{table*}[t]
\centering
\caption{Representative pricing and technical characteristics of major foundation models (prices in USD per million tokens; values are representative and subject to change).}
\label{tab:foundation_models}
\small
\begin{tabularx}{\textwidth}{@{}p{1.65cm}p{2.15cm}p{1.35cm}p{1.1cm}p{1.15cm}p{1.35cm}Y@{}}
\toprule
Provider & Model & Context Window & Input & Output & Cached Input & Primary Strength \\
\midrule
OpenAI &
GPT-5.5 &
1.1M &
\$5.00 &
\$30.00 &
\$0.50 &
Frontier reasoning and coding \\

OpenAI &
GPT-5.4 &
1.1M &
\$2.50 &
\$15.00 &
\$0.25 &
General-purpose reasoning \\

OpenAI &
GPT-5.4 Mini &
1.1M &
\$0.75 &
\$4.50 &
\$0.075 &
Agents and subagents \\

Anthropic &
Claude Opus 4.8 &
1.0M &
\$5.00 &
\$25.00 &
Tiered &
Highest-end reasoning \\

Anthropic &
Claude Sonnet 4.6 &
1.0M &
\$3.00 &
\$15.00 &
Tiered &
Coding and enterprise agents \\

Anthropic &
Claude Haiku 4.5 &
1.0M &
\$1.00 &
\$5.00 &
Tiered &
Low-cost inference \\

Google &
Gemini 3.1 Pro Preview &
1.0M &
\$2--\$4 &
\$12--\$18 &
Context caching &
Long-context reasoning \\

Google &
Gemini 2.5 Flash &
1.0M &
\$0.30 &
\$2.50 &
Context caching &
Cost-efficient hybrid reasoning \\

xAI &
Grok-4.3 &
1.0M &
\$1.25 &
\$2.50 &
\$0.20 &
Tool-oriented API usage \\

DeepSeek &
DeepSeek-V4 &
1.0M &
\$0.44 &
\$0.87 &
-- &
Low-cost frontier model \\

Qwen &
Qwen 3.6 Max &
262K &
\$1.04 &
\$6.24 &
-- &
Multilingual performance \\

GLM &
GLM-5.1 &
203K &
\$0.98 &
\$3.08 &
-- &
Chinese-language applications \\

Llama &
Llama 4 API Offerings &
Varies &
Varies &
Varies &
-- &
Open-weight ecosystem \\

\bottomrule
\end{tabularx}
\end{table*}

Several important patterns emerge from Table~\ref{tab:foundation_models}.
First, output tokens are consistently priced more expensively than input
tokens, with ratios ranging from roughly $2$ for some xAI and DeepSeek
offerings to roughly $6$ or higher for several OpenAI, Google, and Qwen
offerings. Second, providers increasingly differentiate between standard and
cached inputs
\cite{openai_pricing_2026,anthropic_pricing_2026,
google_gemini_pricing_2026,xai_pricing_2026}. OpenAI reports discounted
cached-input rates and batch-processing discounts; Anthropic reports tiered
cache-write and cache-hit prices; Google reports context-caching mechanisms;
and several open-weight or third-party API offerings either omit cached-input
pricing or report it through separate infrastructure layers. Third, pricing
increasingly depends on service mode, including batch processing,
long-context thresholds, priority processing, tool use, data-residency
options, and whether the model is consumed through a hosted API or an
open-weight deployment stack.

The most striking characteristic of contemporary AI pricing models is the
systematic disparity between input and output token costs. Let
$p_I,p_O\in\R_+$ denote the prices of input and output tokens, respectively,
and assume $p_I>0$ when forming price ratios. Across most commercial
offerings, the ratio $\rho=p_O/p_I$ typically exceeds unity and often lies
between four and six. This asymmetry arises because output generation requires repeated
autoregressive decoding operations. While input tokens are processed largely
in parallel during the initial forward pass, output tokens are generated
sequentially through repeated prediction and sampling steps
\cite{vaswani2017attention,brown2020language}. Consequently, each additional
output token imposes both computational and latency costs that exceed those
associated with an additional input token. From an economic perspective,
$\rho$ may be interpreted as a measure of the relative scarcity of generation
capacity. Higher values of $\rho$ encourage concise outputs and discourage
unnecessarily verbose responses.

As context windows continue to expand, providers have also introduced
specialized pricing mechanisms for cached tokens. Let
$T_{\mathrm{cache}}\in\Z_{\ge0}$ denote the number of tokens stored in a
reusable cache. Rather than repeatedly charging the full input price $p_I$,
providers often apply a discounted rate $p_C\in[0,p_I)$ for cache retrieval
operations \cite{openai_pricing_2026,anthropic_pricing_2026,
google_gemini_pricing_2026}. This
practice reflects the observation that retrieving previously computed
representations consumes significantly fewer resources than reprocessing the
same information from scratch. Economically, caching transforms tokens from a
pure flow resource into a partially reusable asset.

The API prices in Table~\ref{tab:foundation_models} should also be distinguished
from subscription and enterprise offerings. Consumer and team products often
bundle model access behind seat-based or monthly plans, while enterprise
contracts may combine seats, usage at API rates, spend controls, service-level
agreements, dedicated support, and platform-specific deployment features
\cite{claude_enterprise2026,google_agent_platform2026}. Hence, public
per-token prices reveal only part of the economic contract: the effective
price paid by an organization also depends on volume commitments, latency
requirements, governance controls, data-residency constraints, and the degree
to which agent platforms bundle orchestration and monitoring into the service.

\subsection{Subscription, Enterprise, and Hybrid Pricing}

Although token pricing dominates API-based services, many providers also offer
subscription-based access. Under this model, users pay a fixed monthly fee in
exchange for access to a specified collection of models and usage limits.
Examples include consumer-facing plans such as ChatGPT Plus and premium
offerings from competing providers. Subscription pricing effectively pools token demand across users. Heavy users
consume more computational resources than their subscription fees directly
cover, whereas light users subsidize excess capacity through underutilization.
From the provider's perspective, subscriptions reduce revenue volatility and
simplify budgeting for customers. From the user's perspective, subscriptions
replace uncertain token expenditures with predictable recurring costs. Enterprise agreements frequently combine multiple pricing mechanisms. A
typical contract may include seat licenses, committed spending thresholds,
discounted token rates, dedicated infrastructure, service-level agreements,
and reserved throughput guarantees. Such arrangements increasingly decouple
pricing from raw token counts and instead align billing with organizational
value creation. Current industry practice remains largely centered on token accounting.
Nevertheless, several trends suggest movement toward more sophisticated
pricing structures. First, long-context reasoning models increasingly
differentiate between standard tokens and reasoning-intensive computation.
Second, agentic systems consume tokens across multiple interconnected model
calls, making direct token accounting less transparent
\cite{yao2023react,schick2023toolformer,yang_zhu2025pact}. Third, enterprise customers
increasingly evaluate AI services according to business outcomes rather than
computational inputs. These developments suggest that token pricing may
eventually coexist with mechanisms based on tasks, outcomes, subscriptions, or
hybrid arrangements.

\subsection{Token-Level Instrumentation and Hidden Reasoning}

The central objective of token instrumentation is to establish mappings between
observed token usage and the underlying computational resources consumed
during inference. Formally, if $T\in\R_+$ denotes total token consumption, one
seeks to estimate functions $g,h,k:\R_+\rightarrow\R_+$ such that
$C(T)=g(T)$, $\mathcal E(T)=h(T)$, and $F(T)=k(T)$, where $C(T)$ denotes
monetary cost, $\mathcal E(T)$ denotes the model-based energy consumption
associated with a workload of size $T$, and $F(T)$ denotes computational
effort measured in floating-point operations (FLOPs). Such mappings provide the foundation for connecting token
accounting to resource allocation, pricing, and economic value.

One of the most challenging aspects of AI tokenomics is the quantification of
hidden reasoning tokens. Advanced reasoning models frequently perform
intermediate computational processes that are only partially exposed through
observable outputs. Recent work on chain-of-thought reasoning,
self-consistency, latent reasoning trajectories, and model interpretability
suggests that substantial computational effort may occur through internal
representations that are not directly visible to users
\cite{wei2022chain,wang2023selfconsistency,
belrose2023eliciting,elhage2021mathematical}.

Let $T_H\in\Z_{\ge0}$ denote the number of hidden reasoning tokens. The true
token footprint of a task follows Eq.~\eqref{eq:token-footprint}, where only
the first four quantities are typically observable. Several approaches have
been proposed for estimating $T_H$. One strategy compares token usage under
direct-answer prompting and explicit chain-of-thought prompting. Another
approach leverages interpretability tools such as the logit lens and tuned
lens, which provide partial visibility into intermediate representational
states \cite{nostalgebraist2020logitlens,belrose2023eliciting}. More
generally, hidden-token estimation may be formulated as an inverse problem in
which observable outputs are used to infer latent computational activity.

\subsection{Computational and Energy Profiling}

Token counts alone provide only a partial characterization of resource
consumption. Two models may process the same number of tokens while consuming
vastly different amounts of computation. Consequently, instrumentation must
also quantify the computational effort associated with token processing. For
a transformer containing $L$ layers and model width $m$, inference cost
scales approximately as in Eq.~\eqref{eq:flops-token}, where $\kappa$ is an
architecture-dependent constant
\cite{vaswani2017attention,kaplan2020scaling,hoffmann2022training}. Modern
profiling frameworks therefore combine analytical models with GPU telemetry,
performance counters, and energy monitors
\cite{dao2022flashattention,reddi2020mlperf}. GPU utilization, memory
bandwidth, execution latency, and throughput can be monitored through tools
such as NVIDIA Management Library (NVML), Data Center GPU Manager (DCGM), and
vendor-specific telemetry interfaces.

The growing scale of AI workloads has elevated energy consumption to a central
economic and environmental concern
\cite{strubell2019energy,wu2022sustainableai}. Let $\bar e\in\R_+$ denote the
average energy expenditure associated with processing a single token. The
token-based energy approximation for a workload containing $T\in\R_+$ tokens
is $\mathcal E_{\mathrm{tok}}(T)=\bar e T$. This relationship motivates a useful performance
metric: energy per token. Measured in Joules per token, this quantity provides
a hardware-independent indicator of inference efficiency and facilitates
comparisons across model architectures and deployment environments
\cite{strubell2019energy,wu2022sustainableai}.

If $\tau>0$ is the inference duration and
$P:[0,\tau]\rightarrow\R_+$ denotes integrable instantaneous power
consumption, the realized energy consumed during the inference session is

\[
E_{\mathrm{sess}}=\int_0^\tau P(s)\,ds.
\]

\noindent The quantity $\mathcal E_{\mathrm{tok}}(T)$ is therefore a token-based
energy model, whereas $E_{\mathrm{sess}}$ is a realized physical measurement.
Dividing $E_{\mathrm{sess}}$ by the number of processed tokens yields an
empirical estimate of energy efficiency that can be compared across hardware
platforms, model families, and operational settings.

\section{Market and Mechanism Design}
\label{sec:markets}

The preceding pricing discussion treats token prices mainly as provider-set
accounting rules. This section broadens the perspective from bilateral
provider-user pricing to market and mechanism design. Once tokens are viewed
as scarce units of computational reasoning capacity, questions of allocation,
congestion, incentives, governance, and strategic interaction naturally arise.

\subsection{Token Markets and Allocation Mechanisms}

The preceding sections established that tokens function simultaneously as
units of information, computation, and pricing. From an economic perspective,
this suggests that tokens may be treated as scarce resources subject to
allocation mechanisms. The study of token markets therefore naturally
intersects with mechanism design, auction theory, cloud-resource allocation,
and algorithmic game theory \cite{myerson1981optimal,nisan2007algorithmic}.
In this setting, a mechanism is not limited to an auction. More broadly, it is
a rule that maps information about demand, supply, priorities, or workflow
needs into token allocations, payments, access rights, or service guarantees.
This broader interpretation is important because AI token allocation appears
at several layers: provider-user pricing, cloud-capacity allocation,
enterprise budget assignment, agent-to-agent delegation, and internal workflow
optimization.

Formally, consider $N\in\N$ users or organizational units indexed by
$\mathcal I=\{1,\ldots,N\}$. Let $\mathcal R_i$ denote the report or bid
space of participant $i$, let $\mathcal R=\prod_{i\in\mathcal I}\mathcal R_i$,
and let $\mathcal F_N(\bar T)=\{a\in\R_+^N:\sum_{i\in\mathcal I}a_i\le\bar T\}$
denote the feasible set of aggregate token allocations. A token-allocation
mechanism may be represented by an allocation rule
$\mathsf X:\mathcal R\rightarrow\mathcal F_N(\bar T)$ and, when monetary
transfers are present, a payment rule
$\mathsf Y:\mathcal R\rightarrow\R_+^N$. The design objective may be welfare
maximization, cost recovery, incentive compatibility, fairness, robustness,
or a combination of these criteria. Fixed posted prices, congestion prices,
auctions, enterprise contracts, and internal budget-allocation rules can
therefore be interpreted as different mechanisms for governing scarce token
capacity.

Let $\bar T\in\R_+$ denote the aggregate supply of computational tokens
available within a platform. Users compete for access to these resources,
generating a market-demand function
$D_{\mathrm{mkt}}:\R_+\rightarrow\R_+$, where $\pi\in\R_+$ denotes the
scalar token price and $D_{\mathrm{mkt}}(\pi)$ denotes aggregate demand at
that price. The resulting allocation problem resembles classical
resource-allocation problems in communication networks, cloud computing, and
electricity markets.

Most current AI services employ fixed token prices. However, fixed pricing may
be inefficient when demand fluctuates substantially over time. Let $H>0$
denote a pricing horizon. Let $D_H,S_H:[0,H]\rightarrow\R_+$ denote
time-dependent demand and supply with $S_H(t)>0$, and let
$\pi_H:[0,H]\rightarrow\R_+$ denote the time-dependent token price. A simple
congestion-pricing rule may be expressed as

\[
\pi_H(t)
=
\pi_0
+
\chi
\left(
\frac{D_H(t)}{S_H(t)}
\right),
\]

\noindent where $\pi_0\in\R_+$ is a base price and $\chi\ge0$ is a congestion
sensitivity parameter. Such pricing mechanisms increase token prices during
periods of high congestion and decrease prices when resources are
underutilized, encouraging more efficient allocation of computational capacity
\cite{kelly1998rate,ghodsi2011dominant}.

Auction mechanisms provide another allocation approach. In an auction
specialization, participant $i\in\mathcal I$ receives allocation
$T_i\in\R_+$ and derives utility $U_i(T_i)$, where
$U_i:\R_+\rightarrow\R$ is nondecreasing and concave. The platform seeks an
allocation maximizing aggregate welfare, $\max_{\{T_i\}}\sum_{i\in\mathcal I}
U_i(T_i)$ subject to $\sum_{i\in\mathcal I}T_i\le\bar T$. Auction-based
mechanisms provide a principled approach for allocating scarce computational
resources while revealing users' willingness to pay
\cite{myerson1981optimal,nisan2007algorithmic}.
The auction formulation is most natural for external markets with strategic
participants. In contrast, an enterprise may use an internal allocation
mechanism in which workflows do not bid in a market but are assigned tokens
according to measured productivity, dependency structure, risk, and business
priority. Section~\ref{sec:workflow} develops precisely this internal
workflow-allocation mechanism. It specializes the general mechanism-design
view above to a planner-mediated setting in which the allocation rule maps a
workflow graph, production functions, a finite budget, and a risk functional
into a feasible token-allocation vector.

\subsection{Tokenized Compute and Multi-Agent Economies}

Recent developments in decentralized infrastructure and distributed computing
have proposed treating computational capacity itself as a tradable asset
\cite{protocol_labs2017filecoin,akash2020decloud}. Under this paradigm,
ownership rights to GPUs, compute clusters, storage resources, or future
inference capacity may be represented through digital tokens. Such proposals
illustrate a broader trend toward the financialization of computational
resources and the emergence of secondary markets for AI infrastructure. If
successful, these systems may improve utilization, liquidity, and access to
computational resources while introducing new questions regarding regulation,
governance, and market stability.

The emergence of autonomous AI agents introduces a second layer of economic
interaction
\cite{yao2023react,schick2023toolformer,park2023generative,chen_zhu2026neuroai,
yang_zhu2026internet_agentic_ai}.
The notion of computational agents exchanging scarce resources has long been
studied in distributed artificial intelligence and computational economics
\cite{wooldridge2009introduction,shoham2008multiagent,zhu_han2026cognitive}.
Let $n\in\N$ and let
$\mathcal A=\{A_1,\ldots,A_n\}$ denote a finite collection of agents. Agent
$A_i$ may allocate tokens to agent $A_j$ in exchange for performing a subtask.
Such interactions create internal token economies governed by incentive
mechanisms, budget constraints, and strategic behavior. As agentic AI systems
become increasingly autonomous, tokens may evolve from a billing unit into an
internal computational currency that coordinates distributed reasoning and
resource allocation.

\subsection{Governance and Open Market-Design Questions}

As token markets evolve, governance and regulatory concerns become
increasingly important. Token-based allocation mechanisms may inadvertently
create barriers to access, favor wealthier participants, or concentrate
computational resources among a small number of actors. Questions of fairness
and access closely parallel earlier concerns in communication networks, cloud
computing, and platform economics
\cite{kelly1998rate,ghodsi2011dominant,acemoglu2021harms,
zhu2024cyberresilience,zhu_basar2024sociotechnical}. Consequently, future AI marketplaces must balance efficiency with broader
societal objectives, including transparency, accessibility, competition, and
equitable access to computational resources. The convergence of AI services,
cloud computing, and digital markets suggests the emergence of a new class of
economic systems in which tokens represent computational resources.
Understanding these systems requires integrating insights from computer
science, economics, operations research, mechanism design, and game theory.
Cyber-resilience and socio-technical-control perspectives further emphasize
that market rules must account for adversarial behavior, bounded rationality,
and feedback between human organizations and AI infrastructure
\cite{zhu2024cyberresilience,zhu_basar2024sociotechnical,
zhu2025gametheory_llm_agentic,albari_zhu2025gestalt_workflows,
zhu_albari2026cyberdeception}.

The study of token markets therefore intersects naturally with algorithmic
game theory, mechanism design, cloud-resource allocation, and digital platform
economics \cite{myerson1981optimal,nisan2007algorithmic,rochet2003platform}.
Future research must address questions concerning equilibrium token prices,
optimal allocation mechanisms, dynamic market design, incentive compatibility,
and the interaction between computational scarcity and economic value. Such
questions constitute the foundation of a broader theory of AI token markets and
provide a natural extension of the emerging field of AI tokenomics.

The next section moves from this market-level perspective to the enterprise
workflow level. Rather than introducing a separate topic, it studies one
concrete mechanism in the taxonomy above: an internal resource-allocation rule
for assigning a finite token budget across interdependent AI workflows.

%%%%%%%%%%%%%%%%%%%%%%%%%%%%%%%%%%%%%%%%%%%%%%%%%%%%%%%%%%%%%%%%%%%%%%
\section{Workflow Tokenomics and Resource Allocation}
\label{sec:workflow}
%%%%%%%%%%%%%%%%%%%%%%%%%%%%%%%%%%%%%%%%%%%%%%%%%%%%%%%%%%%%%%%%%%%%%%

This section develops the workflow-level allocation model that connects
task-dependent token demand to enterprise value. Whereas the previous
sections focused on token definitions, measurement, and pricing, the analysis
below treats tokens as controllable resources that must be distributed across
interdependent AI workflows under an exogenous budget or capacity limit. In
this sense, the total amount of available computation is not the object being
priced; it is the constraint under which the system must operate. The
fundamental planning question is how a finite token budget should be assigned
across workflow components when quality, marginal productivity, downstream
propagation, and risk all depend on the allocation.

Following the mechanism-design perspective in Section~\ref{sec:markets}, the
model below can be interpreted as a concrete enterprise-internal allocation
mechanism: a rule that maps workflow dependencies, token-productivity
functions, budget constraints, and risk exposure into a feasible allocation of
tokens. This viewpoint is consistent with recent work on agentic AI workflows
in cyber deception, where planning, monitoring, deception design, and defensive
response form coupled stages rather than isolated model calls
\cite{albari_zhu2025gestalt_workflows,zhu_albari2026cyberdeception}. The
problem is fundamental because every deployed AI system faces some binding
resource envelope, whether expressed as API spending, latency, energy,
throughput, memory, or context-window capacity. Once that envelope is fixed,
each token allocated to one workflow stage creates an opportunity cost for
the others, and the best allocation cannot be inferred from task-level token
counts alone.

\subsection{From Token Demand to Fixed-Budget Workflow Allocation}

The preceding sections established that tokens constitute the fundamental
resource through which modern AI systems consume computation, memory,
energy, and monetary expenditure. At the task level, token demand was
modeled by the token-demand function in Eq.~\eqref{eq:token-demand}. This
relationship explains how computational requirements emerge from task
characteristics and why token consumption varies substantially across
applications.

Section~\ref{sec:markets} then situated token allocation within the broader
language of mechanisms, including pricing rules, auctions, contracts,
agent-to-agent exchange, and internal allocation rules. The present section
focuses on the last of these mechanisms. It studies a single organization or
platform that controls a finite token budget and must allocate it across
interdependent workflows without relying on an external market-clearing
process. Put differently, pricing and contracts help determine how token
capacity is charged, procured, or shared across agents; after a budget has
been set, the operator still faces the internal workflow-allocation problem
studied here.

While the task-level perspective provides insight into the determinants of
token demand, many practical AI deployments consist of collections of
interconnected workflows operating under common computational and economic
constraints. Examples include RAG, staged reasoning, coding, research support,
cybersecurity, cyber deception, planning, and decision support
\cite{albari_zhu2025gestalt_workflows,zhu2025gametheory_llm_agentic,
zhu_albari2026cyberdeception}. In such environments, the
central challenge is no longer determining how many tokens a single task
requires, but rather how a given token budget should be allocated across
multiple workflows to maximize overall system performance and value. This
allocation problem is a primitive of AI tokenomics: it appears before one can
evaluate return on token spending, compare competing workflow designs, or
decide whether additional capacity is worth purchasing.
The cyber-deception examples are particularly instructive because the value
of an agentic workflow depends on how sensing, hypothesis generation,
deception selection, adversary modeling, and response planning reinforce one
another \cite{albari_zhu2025gestalt_workflows,zhu_albari2026cyberdeception}.
This observation motivates a workflow-level theory of token allocation.
The resulting problem is closely related to classical resource-allocation,
network optimization, workflow scheduling, and utility-maximization
problems studied in economics, operations research, communication
networks, and cloud computing
\cite{kelly1998rate,bertsekas1998network,
topcuoglu2002performance,ghodsi2011dominant,zhu_basar2024sociotechnical,
zhu_han2026cognitive,yang_zhu2026internet_agentic_ai}. The distinguishing feature is that the scarce
resource is computational reasoning capacity measured in tokens.
Conceptually, workflow tokenomics occupies the intermediate layer between
task-level token demand and system-level value creation: task
characteristics determine token demand, token allocations generate workflow
quality, and workflow quality determines system utility.

\subsection{Workflow Graph Representation}

Let $W\in\N$ denote the number of workflows and let
$\Wcal=\{1,\ldots,W\}$ denote the finite workflow index set. Each
$w\in\Wcal$ may represent a retrieval process, a reasoning stage, a planning
module, a verification process, a monitoring function, or any other
structured computational activity.
Dependencies among workflows are represented by a directed graph
$G=(\Wcal,E)$ with $E\subseteq\Wcal\times\Wcal$. An edge $(j,w)\in E$
indicates that the output produced by workflow $j$ serves as an input to
workflow $w$.
For each workflow $w$, define
$\mathrm{Pa}(w)=\{j:(j,w)\in E\}$ as the set of upstream workflows and
$\mathrm{Ch}(w)=\{k:(w,k)\in E\}$ as the set of downstream workflows.
The graph representation captures a broad range of workflow structures.
For example, a retrieval workflow may provide contextual information to a
reasoning workflow, whose output is subsequently passed to a planning
workflow and then to an execution or monitoring workflow.
In agentic cyber-deception workflows, a similar graph can encode how
monitoring, attack inference, deception generation, and defensive control
depend on one another, as in the gestalt game-theoretic workflow models of
\cite{albari_zhu2025gestalt_workflows,zhu_albari2026cyberdeception}.
For analytical tractability, we initially assume that $G$ is a directed
acyclic graph with no self-loops. This assumption permits workflows to be
evaluated in topological order and is consistent with many practical AI
pipelines.
Feedback systems may be represented through temporal unfolding over a
finite horizon, thereby producing an equivalent acyclic workflow graph.

\subsection{Token Allocation and Workflow Quality}

Each workflow receives a token allocation $T_w\in\R_+$. Consistent with the
token taxonomy in Eq.~\eqref{eq:token-footprint}, this allocation may be
decomposed as
$T_w=T_w^{I}+T_w^{C}+T_w^{R}+T_w^{O}+T_w^{H}$, where $T_w^{I}$, $T_w^{C}$,
$T_w^{R}$, $T_w^{O}$, and $T_w^{H}$ denote input, context, retrieval,
output, and hidden reasoning tokens, respectively.
The aggregate token-allocation vector is
$T=(T_1,\ldots,T_W)\in\R_+^W$. Suppose that the system operates under a finite
token budget $B\in\R_{++}$. The feasible allocation set is
$\mathcal F(B)=\{T\in\R_+^W:\sum_{w\in\Wcal}T_w\le B\}$. This budget may
represent an API expenditure limit, a computational capacity constraint, a
latency budget, an energy budget, or a combination of these considerations.
Associated with each workflow is a quality space
$\mathcal Q_w\subseteq\R_+$. The quality variable $q_w\in\mathcal Q_w$ may
represent predictive accuracy, reliability, robustness, information quality,
task-completion probability, or expected operational effectiveness. Let
$\mathcal Q=\prod_{w\in\Wcal}\mathcal Q_w$ denote the system quality space.

Workflow quality is modeled through a production function
$\phi_w:\R_+\times\prod_{j\in\mathrm{Pa}(w)}\mathcal Q_j\rightarrow\mathcal Q_w$:

\begin{equation}
q_w
=
\phi_w
\!\left(
T_w,
\{q_j\}_{j\in\mathrm{Pa}(w)}
\right).
\label{eq:workflow-quality}
\end{equation}

\noindent This formulation generalizes classical production functions from
economics to workflow-based AI systems. The token allocation serves as a
computational input, while upstream workflow qualities influence
downstream performance.
Empirical evidence from scaling laws, retrieval-augmented generation, and
reasoning models suggests several regularities. First, workflow quality
typically improves as token allocation increases
\cite{kaplan2020scaling,hoffmann2022training}. Second, marginal gains
generally diminish as token expenditure increases. Third, downstream
workflow quality depends critically on the quality of upstream
information sources \cite{lewis2020rag}. Consequently, in the continuous
relaxation we assume that $\phi_w$ is continuously differentiable,
nondecreasing and concave in $T_w$, and nondecreasing in each upstream
quality argument. If $\mathrm{Pa}(w)=\emptyset$, the product over upstream
quality spaces is interpreted as a singleton.

A representative specification is

\begin{equation}
q_w
=
\left(
1-e^{-\beta_w T_w}
\right)
g_w
\!\left(
\{q_j\}_{j\in\mathrm{Pa}(w)}
\right),
\label{eq:saturating-quality}
\end{equation}

\noindent where $\beta_w>0$ measures token efficiency and
$g_w:\prod_{j\in\mathrm{Pa}(w)}\mathcal Q_j\rightarrow\R_+$ is a
nonnegative aggregation function for upstream workflow influence.

Because $G$ is acyclic, Eq.~\eqref{eq:workflow-quality} defines a unique
quality vector $q(T)=(q_1(T),\ldots,q_W(T))\in\mathcal Q$ recursively in
topological order for each $T\in\mathcal F(B)$. Thus the workflow network
induces a nonlinear mapping $q:\mathcal F(B)\rightarrow\mathcal Q$. Because
workflow qualities are coupled through the graph structure, this mapping is
generally nonseparable. Consequently, the value of tokens allocated to a
workflow cannot be assessed solely through local performance improvements.

\subsection{System Utility and Network Token Productivity}

The objective of token allocation is to maximize the value generated by
the workflow system. Let $U:\mathcal Q\rightarrow\R$ be a continuously
differentiable system-utility function. The induced value of a token
allocation is

\begin{equation}
\mathcal V(T)
=
U\!\big(q(T)\big)
\label{eq:system-utility}
\end{equation}

\noindent for $T\in\mathcal F(B)$.

A useful benchmark specification is
$\mathcal V(T)=\sum_{w\in\Wcal}b_w q_w(T)$, where
$b_w>0$ measures the relative importance of workflow $w$.

Unlike conventional token accounting, the utility in
Eq.~\eqref{eq:system-utility} depends on workflow interactions. The value of
a token allocated to workflow $w$ depends not only on local quality
improvements but also on the influence of workflow $w$ on downstream
workflows throughout the network.

Following the network-utility literature
\cite{kelly1998rate,bertsekas1998network}, define the adjoint value
variable recursively in reverse topological order by

\begin{equation}
\mu_w
=
\frac{\partial U}{\partial q_w}
+
\sum_{k\in\mathrm{Ch}(w)}
\mu_k
\frac{\partial q_k}{\partial q_w}.
\label{eq:adjoint-value}
\end{equation}

\noindent The quantity $\mu_w$ measures the total marginal value generated by an
incremental improvement in workflow $w$, including both direct and
indirect downstream effects.

Applying the chain rule yields

\begin{equation}
\frac{\partial \mathcal V}{\partial T_w}
=
\mu_w
\frac{\partial \phi_w}{\partial T_w}.
\label{eq:marginal-productivity}
\end{equation}

\noindent This quantity represents the marginal productivity of tokens allocated to
workflow $w$.

The expression generalizes the earlier task-level productivity measure
$\eta=U/T$ by incorporating workflow position and dependency structure. Two
workflows receiving identical token allocations may therefore exhibit
substantially different marginal values depending upon their locations
within the workflow graph.

\subsection{Risk-Aware Workflow Allocation}

The preceding formulation assumes that workflow quality and system value are
deterministic functions of token allocations. In practice, however, AI
workflows operate under substantial uncertainty. Retrieval modules may fail
to recover relevant information, reasoning quality may vary across model
executions, tool invocations may be unreliable, and autonomous agents may
propagate errors throughout downstream workflows. Furthermore, failures in
AI-enabled systems may generate operational losses, cybersecurity incidents,
regulatory penalties, or legal liabilities
\cite{zhu2026insurance_agentic_ai}.

Consequently, token allocation influences not only expected utility but also
the uncertainty, fragility, resilience, and liability exposure of the
workflow network. Two token allocations may generate identical expected
value while exhibiting substantially different operational risk profiles.
For example, excessive concentration of tokens in a small number of
workflows may create infrastructure bottlenecks, increase congestion in
shared computational resources, amplify error propagation, and reduce
overall system resilience.

Let $(\Omega,\mathcal G,\mathbb P)$ be a probability space. Its outcomes
represent stochastic inference, retrieval variability, adversarial inputs,
model misspecification, and tool uncertainty. For each $w\in\Wcal$, let
$\widetilde\phi_w$ be a measurable
stochastic production mapping with domain equal to the product of the token
space $\R_+$, the upstream quality space
$\prod_{j\in\mathrm{Pa}(w)}\mathcal Q_j$, and the uncertainty space
$\Omega$, and with codomain $\mathcal Q_w$. For any
$T\in\mathcal F(B)$ and $\omega\in\Omega$, the realized quality vector
$\widetilde q(T,\omega)\in\mathcal Q$ is defined recursively by applying
$\widetilde\phi_w$ to $T_w$, the upstream realized qualities, and
$\omega$. When the expectation exists, the deterministic quality used in
the preceding subsections can be interpreted as
$q_w(T)=\E[\widetilde q_w(T,\omega)]$.

Let $\mathcal O\subseteq\R^W$ be an open set containing $\mathcal F(B)$. A
risk functional is a differentiable mapping $\Phi:\mathcal O\to\R_+$ that
assigns each feasible token-allocation vector $T=(T_1,\ldots,T_W)$ a
nonnegative measure of uncertainty, congestion, fragility, and liability
exposure. The formulation is \emph{risk-aware} because the planner
maximizes the risk-adjusted objective $\mathcal V(T)-\zeta\Phi(T)$ rather
than workflow value alone. Here $\zeta\ge0$ is a risk-awareness parameter:
$\zeta=0$ gives a purely value-maximizing allocation, whereas larger values
of $\zeta$ place greater weight on robustness, resilience, and liability
mitigation.

Four sources of risk are particularly relevant in workflow tokenomics.
First, even under fixed token allocations, workflow outcomes may vary
because of stochastic model behavior, imperfect retrieval, environmental
uncertainty, or adversarial manipulation. Let
$\sigma_w^2:\R_+\to\R_+$ denote a differentiable local variance penalty for
workflow $w$, for example
$\sigma_w^2(T_w)=\operatorname{Var}[\widetilde\phi_w(T_w,
\{q_j(T)\}_{j\in\mathrm{Pa}(w)},\omega)]$. Additional tokens devoted to
retrieval, verification, monitoring, or reasoning may reduce this term and
improve robustness. Second, AI workflows compete for common resources such
as GPU memory, inference throughput, API quotas, context-window capacity,
storage, and latency budgets. Let $\mathcal R$ be a finite set of shared
resource classes, let $a_w^r\ge0$ be the load placed on resource class $r$
per token allocated to workflow $w$, and define the aggregate load as
$x_r(T)=\sum_{w\in\Wcal}a_w^rT_w$. Congestion is represented by
differentiable, nondecreasing functions $\psi_r:\R_+\to\R_+$. Third,
workflow networks create interdependencies through which upstream errors
may propagate downstream. For each edge $(j,w)\in E$, let
$\Sigma_j(T)=\operatorname{Var}[\widetilde q_j(T,\omega)]$ and let
$\kappa_{jw}\ge0$ weight the propagation risk along that edge. The
sensitivity $\partial q_w/\partial q_j$, evaluated at the baseline quality
vector $q(T)$, measures how strongly workflow $w$ depends on the quality of
workflow $j$. Fourth, in safety-critical and enterprise environments,
workflow failures may generate economic losses through cybersecurity
incidents, erroneous legal recommendations, operational disruptions,
regulatory violations, or unsafe autonomous decisions. Let
$\Lambda:\mathcal O\to\R_+$ be a differentiable expected-liability mapping.

Combining these effects yields the representative risk functional

\begin{equation}
\Phi(T)
=
\sum_{w\in\Wcal}
\sigma_w^2(T_w)
+
\sum_{r\in\mathcal R}
\psi_r\bigl(x_r(T)\bigr)
+
\sum_{(j,w)\in E}
\kappa_{jw}
\Sigma_j(T)
\left(
\frac{\partial q_w}{\partial q_j}
\right)^2
+
\Lambda(T)
\label{eq:risk-functional}
\end{equation}

\noindent for $T\in\mathcal F(B)$. The first term captures local
workflow-quality uncertainty, the second shared-resource congestion, the
third dependency-driven error propagation, and the fourth liability
exposure. Other specifications of $\Phi$ are possible; Eq.~\eqref{eq:risk-functional}
is intended as a tractable model that makes the main risk channels explicit.

The workflow-allocation problem becomes

\begin{equation}
\max_{T\in\mathcal F(B)}
\left\{
\mathcal V(T)-\zeta\Phi(T)
\right\}.
\label{eq:risk-aware-allocation}
\end{equation}

\noindent This is a risk-aware, rather than risk-neutral, optimization
problem. Allocations depend on both marginal value and marginal exposure to
uncertainty, congestion, propagation, and liability risk.

\subsection{Optimality Conditions and Economic Interpretation}

The Lagrangian associated with problem
\eqref{eq:risk-aware-allocation} is

\[
\mathcal L
=
\mathcal V(T)
-
\zeta\Phi(T)
+
\lambda
\left(
B-\sum_{w\in\Wcal}T_w
\right).
\]

At an interior optimum, the first-order conditions imply

\begin{equation}
\frac{\partial \mathcal V}{\partial T_w}
-
\zeta
\frac{\partial \Phi}{\partial T_w}
=
\lambda,
\qquad
w\in\Wcal.
\label{eq:first-order-condition}
\end{equation}

Substituting Eq.~\eqref{eq:marginal-productivity} into
Eq.~\eqref{eq:first-order-condition} yields

\begin{equation}
\mu_w
\frac{\partial \phi_w}{\partial T_w}
-
\zeta
\frac{\partial \Phi}{\partial T_w}
=
\lambda.
\label{eq:risk-adjusted-foc}
\end{equation}

Equation~\eqref{eq:risk-adjusted-foc} equates the risk-adjusted marginal
productivity of tokens across workflows. The first term,
$\mu_w \frac{\partial \phi_w}{\partial T_w}$, measures the incremental value
generated by allocating an additional token to workflow $w$, including both
direct and downstream effects. The second term,
$\zeta \frac{\partial \Phi}{\partial T_w}$, measures the incremental
uncertainty, congestion, fragility, and liability introduced by that
allocation.

The multiplier $\lambda$ represents the shadow value of an additional token
and therefore quantifies the marginal value of computational reasoning
capacity under the prevailing risk preferences of the organization.
Workflows whose risk-adjusted marginal productivity exceeds $\lambda$
should receive additional tokens, whereas workflows whose risk-adjusted
contribution falls below $\lambda$ should receive fewer resources.

This result extends AI tokenomics beyond cost accounting and pricing.
Tokens become allocable computational assets whose value depends not only on
their contribution to workflow quality, but also on their influence on
uncertainty, resilience, and systemic risk. As AI workflows become
increasingly autonomous, interconnected, and safety-critical, optimal token
allocation requires balancing productivity against robustness throughout the
entire workflow ecosystem.

%%%%%%%%%%%%%%%%%%%%%%%%%%%%%%%%%%%%%%%%%%%%%%%%%%%%%%%%%%%%%%%%%%%%%%
\section{Case Studies: Workflow Tokenomics in Practice}
\label{sec:case-studies}
%%%%%%%%%%%%%%%%%%%%%%%%%%%%%%%%%%%%%%%%%%%%%%%%%%%%%%%%%%%%%%%%%%%%%%

\begin{figure}[t]
\centering
\includegraphics[width=\textwidth]{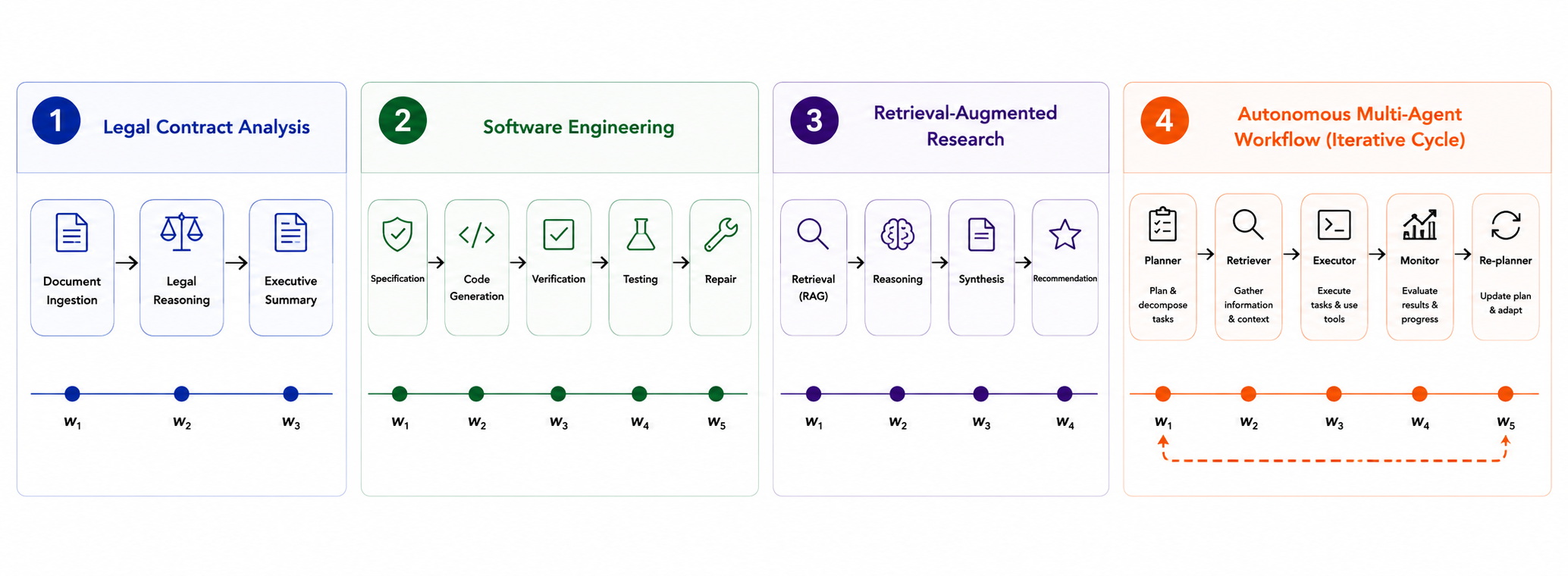}
\caption{Representative workflow structures for the Section~\ref{sec:case-studies}
case studies. From left to right, the panels show legal contract analysis,
software engineering, retrieval-augmented research, and autonomous
multi-agent workflows; the labels $w_1,\ldots,w_W$ identify the workflow
stages used in the allocation calculations.}
\label{fig:workflow-case-studies}
\end{figure}

The preceding sections developed a theory of workflow tokenomics in which
tokens serve simultaneously as units of information representation,
computation, memory consumption, energy expenditure, and economic allocation.
At the task level, token demand follows Eq.~\eqref{eq:token-demand}; at the
workflow level, token allocations are represented by $T=(T_1,\ldots,T_W)$ and
linked to workflow quality through Eq.~\eqref{eq:workflow-quality}. This
section provides numerical case studies showing how token expenditures,
workflow quality, marginal productivity, and economic value interact.
Throughout the case studies, the token cost of a workflow is approximated by
$C=p_I(T^I+T^C+T^R)+p_OT^O+p_HT^H$, where $p_I,p_O,p_H\in\R_+$ denote the
effective prices of input, output, and hidden reasoning tokens. When hidden
reasoning tokens are not separately priced, one may set $p_H=p_O$ or
interpret $p_H$ as the provider's internal marginal inference cost. For
illustration, suppose $p_I=\$2/10^6$, $p_O=\$10/10^6$, and $p_H=\$10/10^6$.
These values are not intended as fixed market prices; they provide a common
numerical basis for comparing token allocation across workflows.
Figure~\ref{fig:workflow-case-studies} summarizes the four workflow
templates used in the case studies and highlights the stage-level variables
$w_1,\ldots,w_W$ that later enter the allocation model.

%%%%%%%%%%%%%%%%%%%%%%%%%%%%%%%%%%%%%%%%%%%%%%%%%%%%%%%%%%%%%%%%%%%%%%
\subsection{Case Study I: Legal Contract Analysis}
%%%%%%%%%%%%%%%%%%%%%%%%%%%%%%%%%%%%%%%%%%%%%%%%%%%%%%%%%%%%%%%%%%%%%%

Corresponding to the first panel of
Figure~\ref{fig:workflow-case-studies}, consider a legal-analysis workflow
with three stages: document ingestion
($w_1$), legal reasoning ($w_2$), and executive summary ($w_3$). Suppose the
portfolio contains ten contracts, each averaging 50 pages, and that one page
corresponds to approximately 200 tokens. Document ingestion therefore requires
$10\times50\times200=100{,}000$ input tokens, so $T_{w_1}^I=100{,}000$. The
reasoning stage performs clause extraction, obligation analysis, risk
identification, compliance checking, and cross-document comparison. If hidden
reasoning consumes between 50\% and 150\% of the ingested document length,
then $T_{w_2}^H\in[50{,}000,150{,}000]$. The final executive summary and risk
report require $T_{w_3}^O=1{,}000$, so
$T_{\mathrm{tot}}=100{,}000+[50{,}000,150{,}000]+1{,}000
\in[151{,}000,251{,}000]$.
Using the illustrative prices, the corresponding cost is computed from input,
hidden reasoning, and output charges, giving $C_{\min}=\$0.71$ and
$C_{\max}=\$1.71$. Thus, even though visible output is only 1,000 tokens,
most cost arises from hidden reasoning, which accounts for approximately
33.1\% to 59.8\% of the total footprint. Let workflow quality be modeled by
$q_w(T_w)=1-\exp(-\beta_wT_w)$. With
$(\beta_1,\beta_2,\beta_3)=(2\times10^{-5},1.5\times10^{-5},10^{-3})$ and
baseline allocation $(T_1,T_2,T_3)=(100{,}000,100{,}000,1{,}000)$, the
qualities are $(q_1,q_2,q_3)=(0.865,0.777,0.632)$. For utility
$\mathcal V=0.1q_1+0.7q_2+0.2q_3$, this gives $\mathcal V=0.7568$.

The marginal quality gain from an additional token is
$dq_w/dT_w=\beta_we^{-\beta_wT_w}$. Under the baseline allocation, the
marginal utilities are $2.71\times10^{-7}$ for ingestion,
$2.34\times10^{-6}$ for reasoning, and $7.36\times10^{-5}$ for the summary.
The high marginal value of summary tokens reflects the small baseline
allocation to the summary stage. However, once the executive summary reaches
a sufficient length, additional summarization tokens saturate rapidly. The
reasoning stage remains economically important because its quality receives
the largest utility weight and propagates downstream into the final legal
recommendation.

%%%%%%%%%%%%%%%%%%%%%%%%%%%%%%%%%%%%%%%%%%%%%%%%%%%%%%%%%%%%%%%%%%%%%%
\subsection{Case Study II: Software Engineering Workflow}
%%%%%%%%%%%%%%%%%%%%%%%%%%%%%%%%%%%%%%%%%%%%%%%%%%%%%%%%%%%%%%%%%%%%%%

The second panel of Figure~\ref{fig:workflow-case-studies} represents the
software-engineering workflow. It has five stages: specification
($w_1$), code generation ($w_2$), verification ($w_3$), testing ($w_4$), and
repair ($w_5$). Suppose $T_{w_1}^I=500$ and $T_{w_2}^O=5{,}000$, while
verification, testing, and repair consume hidden reasoning tokens satisfying
$T_{w_3}^H+T_{w_4}^H+T_{w_5}^H\in[20{,}000,100{,}000]$. Then
$T_{\mathrm{tot}}=500+5{,}000+[20{,}000,100{,}000]\in[25{,}500,105{,}500]$.
Using the illustrative prices, $C_{\min}=0.001+0.05+0.20=\$0.251$ and
$C_{\max}=0.001+0.05+1.00=\$1.051$.
Assume a baseline allocation
$(T_2,T_3,T_4,T_5)=(5{,}000,25{,}000,25{,}000,10{,}000)$ and
$q_w(T_w)=1-e^{-\beta_wT_w}$ with $\beta_2=2\times10^{-4}$,
$\beta_3=\beta_4=8\times10^{-5}$, and $\beta_5=10^{-4}$. The resulting stage
qualities are $q_2=0.632$, $q_3=0.865$, $q_4=0.865$, and $q_5=0.632$. With
$\mathcal V=0.15q_2+0.35q_3+0.35q_4+0.15q_5$, the baseline utility is
$\mathcal V=0.795$.

Now compare two allocations with the same total token budget of 65,000
tokens. Allocation A assigns 15,000 tokens each to code generation,
verification, and testing, and 20,000 tokens to repair; it produces
$(q_2,q_3,q_4,q_5)=(0.950,0.699,0.699,0.865)$ and $\mathcal V_A=0.761$.
Allocation B assigns 5,000 tokens to code generation, 25,000 each to
verification and testing, and 10,000 to repair; it produces
$(q_2,q_3,q_4,q_5)=(0.632,0.865,0.865,0.632)$ and $\mathcal V_B=0.795$.
Although Allocation A devotes more tokens to code generation and repair,
Allocation B yields higher utility because verification and testing receive
larger weights: $\mathcal V_B-\mathcal V_A=0.034$, a relative improvement of
$0.034/0.761\approx4.47\%$. This illustrates a key workflow-tokenomics point:
allocating tokens to the largest or most visible stage is not necessarily
optimal. Verification and testing may dominate marginal productivity because
failures propagate downstream into debugging, deployment delay, security
risk, and user harm.

%%%%%%%%%%%%%%%%%%%%%%%%%%%%%%%%%%%%%%%%%%%%%%%%%%%%%%%%%%%%%%%%%%%%%%
\subsection{Case Study III: Retrieval-Augmented Research Workflow}
%%%%%%%%%%%%%%%%%%%%%%%%%%%%%%%%%%%%%%%%%%%%%%%%%%%%%%%%%%%%%%%%%%%%%%

The third panel of Figure~\ref{fig:workflow-case-studies} illustrates a
retrieval-augmented research workflow with retrieval ($w_1$),
reasoning ($w_2$), synthesis ($w_3$), and recommendation ($w_4$). Suppose the
user query contains $T^I=500$ tokens, the retrieval stage accesses fifty
documents of 1,000 tokens each so that $T^R=50{,}000$, the final synthesis and
recommendation stages produce $T^O=2{,}000$ output tokens, and hidden
reasoning consumes $T^H\in[20{,}000,80{,}000]$. Therefore
$T_{\mathrm{tot}}=500+50{,}000+2{,}000+[20{,}000,80{,}000]
\in[72{,}500,132{,}500]$. The corresponding cost range is
$C_{\min}=0.101+0.020+0.200=\$0.321$ to
$C_{\max}=0.101+0.020+0.800=\$0.921$.

Suppose retrieval quality follows $q_R(T_R)=1-e^{-\beta_RT_R}$ with
$\beta_R=8\times10^{-5}$. At $T_R=50{,}000$, $q_R=1-e^{-4}=0.982$ and
$\partial q_R/\partial T_R=8\times10^{-5}e^{-4}=1.47\times10^{-6}$. Suppose
reasoning quality follows $q_H(T_H)=1-e^{-\beta_HT_H}$ with
$\beta_H=3\times10^{-5}$. At $T_H=40{,}000$, $q_H=1-e^{-1.2}=0.699$ and
$\partial q_H/\partial T_H=3\times10^{-5}e^{-1.2}=9.04\times10^{-6}$. Thus,
at these allocation levels,
$(\partial q_H/\partial T_H)/(\partial q_R/\partial T_R)
=(9.04\times10^{-6})/(1.47\times10^{-6})\approx6.15$. An additional
reasoning token produces over six times the marginal quality gain of an
additional retrieval token. This calculation formalizes the
diminishing-returns claim: once retrieval has already accessed a sufficiently
large and relevant corpus, marginal value shifts toward reasoning, synthesis,
and judgment.

%%%%%%%%%%%%%%%%%%%%%%%%%%%%%%%%%%%%%%%%%%%%%%%%%%%%%%%%%%%%%%%%%%%%%%
\subsection{Case Study IV: Autonomous Multi-Agent Workflow}
%%%%%%%%%%%%%%%%%%%%%%%%%%%%%%%%%%%%%%%%%%%%%%%%%%%%%%%%%%%%%%%%%%%%%%

The fourth panel of Figure~\ref{fig:workflow-case-studies} depicts an
autonomous multi-agent workflow consisting of a planner, retriever, executor,
monitor, and replanner. Suppose one planning cycle requires
$T_P=10{,}000$, $T_R=50{,}000$, $T_E=100{,}000$, $T_M=20{,}000$, and
$T_{RP}=30{,}000$. The total token expenditure is
$T_{\mathrm{tot}}=210{,}000$, with shares
$s_P=4.76\%$, $s_R=23.81\%$, $s_E=47.62\%$, $s_M=9.52\%$, and
$s_{RP}=14.29\%$. Although execution consumes the largest token share, it
need not have the largest marginal value.
To see this, suppose system utility depends on the final replanning quality
$q_{RP}$, and each stage propagates quality multiplicatively so that
$q_{RP}=q_Pq_Rq_Eq_M$. The marginal contribution of planning is
$\partial q_{RP}/\partial q_P=q_Rq_Eq_M$, while the marginal contribution of
execution is $\partial q_{RP}/\partial q_E=q_Pq_Rq_M$. If
$(q_P,q_R,q_E,q_M)=(0.70,0.90,0.85,0.80)$, then
$\partial q_{RP}/\partial q_P=0.90(0.85)(0.80)=0.612$, whereas
$\partial q_{RP}/\partial q_E=0.70(0.90)(0.80)=0.504$. Thus, an improvement
in planning quality has a larger downstream effect than an equivalent
improvement in execution quality: $0.612/0.504=1.214$. Planning has
approximately 21.4\% greater downstream marginal impact despite using only
4.76\% of the tokens. This example demonstrates why adjoint values matter.
The value of a workflow is not determined only by the number of tokens it
consumes; it is determined by its position in the dependency graph and the
extent to which quality improvements propagate downstream.

%%%%%%%%%%%%%%%%%%%%%%%%%%%%%%%%%%%%%%%%%%%%%%%%%%%%%%%%%%%%%%%%%%%%%%
\subsection{Comparative Lessons}
%%%%%%%%%%%%%%%%%%%%%%%%%%%%%%%%%%%%%%%%%%%%%%%%%%%%%%%%%%%%%%%%%%%%%%

The four case studies, visualized in
Figure~\ref{fig:workflow-case-studies}, demonstrate that token expenditure,
marginal value, and economic utility can diverge substantially. The totals in
Table~\ref{tab:workflow_cases_expanded} are computed directly from the
scenario assumptions stated above; they are intended as internally consistent
workload calculations rather than provider-specific empirical averages.

\begin{table}[ht]
\centering
\small
\caption{Expanded workflow tokenomics comparison.}
\label{tab:workflow_cases_expanded}
\begin{tabularx}{\textwidth}{@{}p{2.5cm}p{2.2cm}p{2.4cm}p{2.6cm}Y@{}}
\toprule
Application &
Typical Tokens &
Largest Token Component &
Dominant Marginal-Value Component &
Main Allocation Lesson\\
\midrule

Legal Analysis &
$151$K--$251$K &
Hidden reasoning &
Legal reasoning / risk analysis &
Visible output is small; most value comes from reasoning quality.\\

Software Engineering &
$25.5$K--$105.5$K &
Verification, testing, repair &
Verification and testing &
Quality assurance tokens may outperform code-generation tokens.\\

Research Assistance &
$72.5$K--$132.5$K &
Retrieval and hidden reasoning &
Reasoning after retrieval saturation &
Once retrieval quality saturates, additional reasoning dominates.\\

Multi-Agent Systems &
$210$K per cycle, potentially $1$M+ over repeated cycles &
Execution &
Planning and coordination &
Upstream planning can have higher marginal value than large execution stages.\\

\bottomrule
\end{tabularx}
\end{table}

A common pattern emerges. The stage consuming the most tokens is not
necessarily the stage with the highest marginal productivity. Formally, for
two workflows $i$ and $j$, it is possible that $T_i>T_j$ while
$\partial\mathcal V/\partial T_i<\partial\mathcal V/\partial T_j$. This
inequality is the central economic lesson of workflow tokenomics: allocation
should follow risk-adjusted marginal value, not raw token volume.
Several quantitative lessons follow. First, visible output tokens can be a
small fraction of total token expenditure: in legal analysis, the output
report may require only 1,000 tokens while the total footprint ranges from
151,000 to 251,000 tokens. Second, hidden reasoning can dominate cost: in the
legal case, hidden reasoning accounts for approximately 33.1\% to 59.8\% of
total token consumption, and in software engineering, verification, testing,
and repair can consume four to twenty times as many tokens as code
generation. Third, marginal productivity depends on saturation: in the
research-assistance case, retrieval quality at 50,000 tokens is already
approximately 0.982 under the assumed production function, implying very low
marginal retrieval value. Fourth, workflow position matters: in the
multi-agent case, planning consumes less than 5\% of total tokens but can
have greater downstream marginal impact than execution because planning
quality affects every subsequent stage.

Consequently, the practical objective is not to minimize token use in
isolation. The objective is to solve the risk-aware allocation problem in
Eq.~\eqref{eq:risk-aware-allocation}, with feasible set
$\mathcal F(B)=\{T\in\R_+^W:\sum_{w\in\Wcal}T_w\le B\}$, so that tokens are
assigned to workflows with the highest risk-adjusted marginal value,
satisfying
$\mu_w(\partial \phi_w/\partial T_w)-\zeta(\partial\Phi/\partial T_w)=\lambda$
at an interior optimum. This condition states that every workflow receives
tokens until its risk-adjusted marginal productivity equals the shadow value
of the token budget.

\section{Conclusion and Future Research Directions}
\label{sec:conclusion}

The emergence of foundation models has transformed tokens from a technical
implementation detail into a fundamental economic unit of artificial
intelligence. Tokens now serve simultaneously as units of information
representation, computation, memory utilization, energy expenditure,
resource allocation, and monetary exchange. As a result, understanding the
generation, consumption, pricing, and allocation of tokens has become
essential for the efficient design and operation of modern AI systems.

This paper developed a unified framework for \emph{AI Tokenomics}, defined as
the study of how tokens are generated, consumed, valued, allocated, and
optimized within AI ecosystems. The framework integrates technical,
economic, and operational perspectives into a common analytical structure.
At the technical level, tokens were characterized as computational objects
whose processing determines inference cost, memory requirements, latency,
and energy consumption. At the economic level, tokens were interpreted as
scarce resources whose allocation influences workflow quality, enterprise
value, and organizational performance. At the market level, tokens were
viewed as tradable units that may eventually support new forms of pricing,
allocation, and governance mechanisms.

Several key insights emerge from this perspective. First, token consumption
and economic value are fundamentally distinct concepts. Although tokens
provide a convenient accounting unit for measuring computational effort,
their economic significance depends on the value generated by the tasks they
enable. Second, token demand is highly heterogeneous across applications and
depends upon task difficulty, context size, uncertainty, retrieval
requirements, and workflow structure. Third, enterprise AI systems should be
understood as networks of interdependent workflows in which token
allocations influence both local performance and downstream organizational
outcomes. Finally, efficient token allocation requires balancing
computational cost, productivity, uncertainty, and risk across the entire
workflow ecosystem.

The framework developed herein suggests that AI tokenomics may become a
foundational discipline at the intersection of artificial intelligence,
economics, operations research, and systems engineering. Much as classical
economics studies the allocation of scarce resources and communication
theory studies the allocation of bandwidth, AI tokenomics studies the
allocation of computational reasoning capacity.
Despite recent advances, the field remains in its infancy. Numerous
fundamental questions remain unresolved and provide fertile opportunities
for future research.

One important direction concerns the measurement of hidden reasoning
activity. Contemporary models often perform substantial internal reasoning
that is not directly observable to users. Eq.~\eqref{eq:token-footprint}
defines total token consumption, where $T_H$ represents hidden reasoning
tokens. Existing measurement systems generally observe only the visible
components. Developing methodologies for estimating and validating hidden
reasoning activity remains a central challenge for both technical and
economic analysis.
Relatedly, a second direction concerns empirical calibration.
Eq.~\eqref{eq:token-demand} connects task characteristics to token demand, but
the precise forms of these relationships remain largely unknown. Large-scale
empirical studies are needed to estimate production elasticities, characterize
scaling behavior, and understand how token requirements vary across domains,
models, and organizational contexts.

A third research direction involves the development of rigorous utility and
productivity metrics. The notion of token productivity, $\eta=U/T$, captures
the value generated per token consumed. However, defining and measuring utility
remains difficult. Future work should investigate methods for quantifying
utility through human evaluation, organizational outcomes, economic return,
decision quality, and task-specific performance metrics. Standardized
productivity measures would make comparisons across models, workflows, and
deployment strategies more reliable.
Another important area concerns enterprise token allocation. This paper
introduced optimization models in which token allocations determine workflow
quality and enterprise value. Future work should extend these formulations to
dynamic settings with time-varying demand, learning agents, adaptive budgets,
and strategic interactions among stakeholders. Connections with stochastic
control, network optimization, and multi-agent reinforcement learning appear
particularly promising.

Market and mechanism design constitute another major frontier. Existing AI
markets primarily rely on fixed per-token pricing schemes. Future systems
may employ dynamic pricing, auction-based allocation, congestion pricing,
subscription hybrids, or outcome-based compensation mechanisms. The study
of such systems raises questions concerning efficiency, incentive
compatibility, market equilibrium, and fairness. Understanding the
conditions under which alternative pricing mechanisms outperform
conventional token billing remains an open research problem.
Closely related is the problem of contract design for token-based and
agentic AI services. In many enterprise deployments, a provider does not
merely sell raw tokens; it offers a service contract that may combine token
quotas, service-level guarantees, latency commitments, auditability
requirements, outcome-based payments, and risk-sharing provisions. Designing
such contracts requires addressing adverse selection, moral hazard,
unobservable hidden reasoning effort, heterogeneous customer value, and
uncertain downstream consequences of AI decisions. Contract-theoretic models
of agentic AI pricing, such as PACT \cite{yang_zhu2025pact}, provide a
promising starting point for studying how token prices, quality guarantees,
and incentive-compatible service agreements should be jointly designed.
The emergence of autonomous agents introduces additional challenges. In
future AI ecosystems, agents may allocate, exchange, and negotiate token
budgets among themselves. Such environments naturally give rise to
multi-agent token economies in which computational resources function as an
internal currency. Designing stable, efficient, and incentive compatible token
exchange protocols is a rich interdisciplinary research area spanning
artificial intelligence, economics, and game theory.

An equally important direction concerns the financialization of
computational resources. Recent proposals involving tokenized compute
assets, tradable GPU capacity, and decentralized AI infrastructure suggest
the possibility of secondary markets for computational resources. While
such systems may improve liquidity and resource utilization, they also raise
questions concerning speculation, market concentration, governance, and
regulation. Understanding the economic implications of treating
computational capacity as a financial asset remains an open challenge.
Finally, the development of public benchmarks and datasets is essential for
the maturation of AI tokenomics as a scientific discipline. Future research
should establish standardized datasets linking tasks, token consumption,
workflow characteristics, computational cost, energy expenditure, and
economic outcomes. Such resources would facilitate reproducible research,
enable empirical validation of theoretical models, and support the
development of evidence-based token allocation policies.

In conclusion, tokens have become the fundamental resource through which
modern AI systems consume computation and generate value. As AI systems
become increasingly autonomous, interconnected, and economically important,
the efficient management of token resources will emerge as a central
technical and organizational challenge. AI tokenomics provides a framework
for addressing this challenge by integrating computation, economics,
measurement, optimization, and market design into a unified theory of
resource allocation for computational intelligence.

\bibliographystyle{abbrv}
\bibliography{refs}

\end{document}